\documentclass{article}

\usepackage{PRIMEarxiv}

\usepackage[utf8]{inputenc} 
\usepackage[T1]{fontenc}    
\usepackage{hyperref}       
\usepackage{url}            
\usepackage{booktabs}       
\usepackage{amsfonts}       
\usepackage{nicefrac}       
\usepackage{microtype}      
\usepackage{lipsum}
\usepackage{xcolor} 
\usepackage{amsmath}
\usepackage{graphicx}
\usepackage{algorithm}
\usepackage{soul}
\usepackage{algpseudocode}
\usepackage{fancyhdr}       
\usepackage{graphicx}       
\graphicspath{{media/}}     
\usepackage{subcaption}
\usepackage{multirow}
\usepackage{arydshln}
\title{SelECT-SQL: Self-correcting ensemble Chain-of-Thought for Text-to-SQL

}

\author{
  Ke Shen, Mayank Kejriwal \\
  Information Sciences Institute \\
  University of Southern California \\
  Marina del Rey, CA 90292\\
  \texttt{\{keshen, kejriwal\}@isi.edu} \\
}

\begin{document}
\maketitle

\begin{abstract}
In recent years,\textit{Text-to-SQL}, the problem of automatically converting questions posed in natural language to formal SQL queries, has emerged as an important problem at the intersection of natural language processing and data management research. Large language models (LLMs) have delivered impressive performance when used in an off-the-shelf performance, but still fall significantly short of expected expert-level performance. Errors are especially probable when a nuanced understanding is needed of database schemas, questions, and SQL clauses to do proper Text-to-SQL conversion. We introduce \textbf{SelECT-SQL}, a novel in-context learning solution that uses an algorithmic combination of chain-of-thought (CoT) prompting, self-correction, and ensemble methods to yield a new state-of-the-art result on challenging Text-to-SQL benchmarks. Specifically, when configured using GPT-3.5-Turbo as the base LLM, SelECT-SQL achieves 84.2\% execution accuracy on the Spider leaderboard's development set, exceeding both the best results of other baseline GPT-3.5-Turbo-based solutions (81.1\%), and the peak performance (83.5\%) of the GPT-4 result reported on the leaderboard.
\end{abstract}

\keywords{Text-to-SQL \and Chain-of-thought \and Large language model \and Self-correction \and Ensemble}

\section{Introduction}\label{introduction}

Natural language interfaces to databases allow non-SQL experts to query relational databases more conveniently. \textit{Text-to-SQL}, which automatically maps natural language questions to SQL queries \cite{deng2022recent,katsogiannis2023survey} has therefore emerged as an important problem, especially due to generative AI. Early Text-to-SQL systems were domain-specific with limited user interaction, often relying on rule-based approaches to parse input questions \cite{popescu2003towards,popescu2004modern,li2014constructing,stratica2005using}. Recent advancements have shifted towards greater domain independence by introducing supervised models trained on various cross-domain datasets \cite{zhong2017seq2sql,yu2018spider}, and transformer-based models fine-tuned with built-in modules and constraints \cite{scholak2021picard,xu2022sead,qi2022rasat,li2023graphix}. Unlike retrieval-augmented generation (RAG) \cite{gao2023retrieval}, which uses transformer-based language models fine-tuned on external knowledge, Text-to-SQL reduces potential hallucinations in domain-specific or knowledge-intensive tasks because the answer is from querying the database rather than being generated directly by a model.

Recent developments in Text-to-SQL use large language models (LLMs) with zero-shot \cite{liu2023comprehensive,dong2023c3} and few-shot prompting \cite{gu2023few,gao2023text}, demonstrating that LLMs can serve as strong baselines with minimal demonstration of questions and schemas and no fine-tuning. Despite their potential, GPT-3.5-Turbo often underperforms compared to fine-tuned models on the widely-used Spider benchmark \cite{yu2018spider}, requiring over 10,000 tokens per query \cite{pourreza2023din} or multiple generation rounds for self-consistency \cite{dong2023c3}. PT-4-based solutions perform better, but are less budget-efficient due to higher token costs \footnote{GPT-4 costs \$30 per 1 million input tokens, whereas GPT-3.5-Turbo costs \$0.50 per 1 million input tokens.}.

Few-shot prompting \cite{brown2020language}, especially when combined with Chain-of-Thought (CoT) \cite{wei2022chain} and self-consistency \cite{wang2022self,dong2023c3}, outperforms previous state-of-the-art methods on complex reasoning tasks \cite{li2023symbolic,zhao2023verify}, achieving high accuracy with limited examples. Gao et al. \cite{gao2023text} emphasized that few-shot settings, including example selection and presentation format, significantly affect GPT-family engines' performance on Text-to-SQL tasks. However, other prompting components that can be applied in in-context learning, such as CoT, remain underexplored in Text-to-SQL tasks.

In this paper, we introduce a novel in-context learning solution, \textit{SELf-correcting Ensemble Chain-of-Thought prompting} (\textbf{SelECT-SQL}\footnote{The code is available at \url{https://github.com/neuripsPublishingResearchCode/SelECT-SQL}}), composed of three key components: automatic CoT, self-correction, and ensemble techniques. These components collaboratively refine the generated query but can also be used independently to enhance the model's performance. The proposed structure-synthesis CoT automatically generates example step-by-step solutions, improving GPT-3.5-Turbo's execution accuracy by 4.1\%. The self-correction component systematically and automatically examines the generated query, and adjusts incorrect generation by adopting manually crafted tips to mitigate the prior bias of GPT models in formal query generation. The ensemble component adopts only few generations as potential solutions, improving openness and accuracy without requiring extensive self-consistency checking.

Our contributions can be summarized as follows: (1) proposing a novel prompting strategy that works with generative LLMs, such as GPT-3.5-Turbo, to achieve better performance on the Text-to-SQL task, (2) exploring better configurations of few-shot prompting with GPT-3.5-Turbo as the base LLM, (3) offering deeper insights into scenarios where GPT fails in Text-to-SQL tasks.  We evaluated GPT-3.5-Turbo with SelECT-SQL on the standard Spider benchmark, although it can be extended for use with other LLMs that support in-context learning. It outperforms the best GPT-3.5-Turbo-based solution on the Spider leaderboard\footnote{https://yale-lily.github.io/spider} by 2.4\% and surpasses the best solution working with the GPT-4 engine (without self-consistency) by 0.7\%. The CoT and self-correction components, which use the most lengthy prompting, use approximately 3,000 tokens, while the ensemble component requires only 4 generations instead of dozens for self-consistency. A detailed error analysis of GPT-3.5-Turbo's failures is also provided to offer insights for further improvement in the Text-to-SQL task.

\section{SelECT-SQL}
\label{methodology}

Considering a target question $q$ expressed in natural language related to a specific database $\mathcal D$, which includes a schema $\mathcal S = \{T, C\}$ representing the included tables $T$ and columns $C$ of $\mathcal D$ and outlining its structure, the Text-to-SQL problem prompts model $\mathcal M$ to generate the optimal query $\mathit y$ which maximizes the possibility:

\begin{equation}\label{equation:ori}
\begin{aligned}
\max\limits_{y, \sigma, \mathit{Q'}} &\mathbb{P}_{\mathcal{M}}(y|\sigma(q, D, \mathit{Q'})), \\ 
& s.t.\quad |\mathit{Q'}| = k, \mathit{Q'} \subset \mathit{Q},
\end{aligned}
\end{equation}

where function $\sigma(\cdot,\cdot, \cdot)$ characterizes the representation of question $q$ based on databased $\mathcal D$  and incorporates $k$ selected examples from the triple set $\mathit Q = {(q_i, \mathcal D_i, \mathit y_i)}$. As $k$ increases from 0, the model transitions from zero-shot learning to in-context learning. Based on the definition provided by Gao et al.\cite{gao2023text}, this representation might incorporate various types of information such as the schema $\mathcal S$, and may be expressed in different formats, including SQL syntax and natural language. With the presented formalism, we introduce the three main components of the proposed SelECT-SQL prompting, starting with the Chain-of-thought component.

\subsection{Chain-of-thought Prompting}

\begin{figure}[ht]
  \centering
  \includegraphics[width=\textwidth]{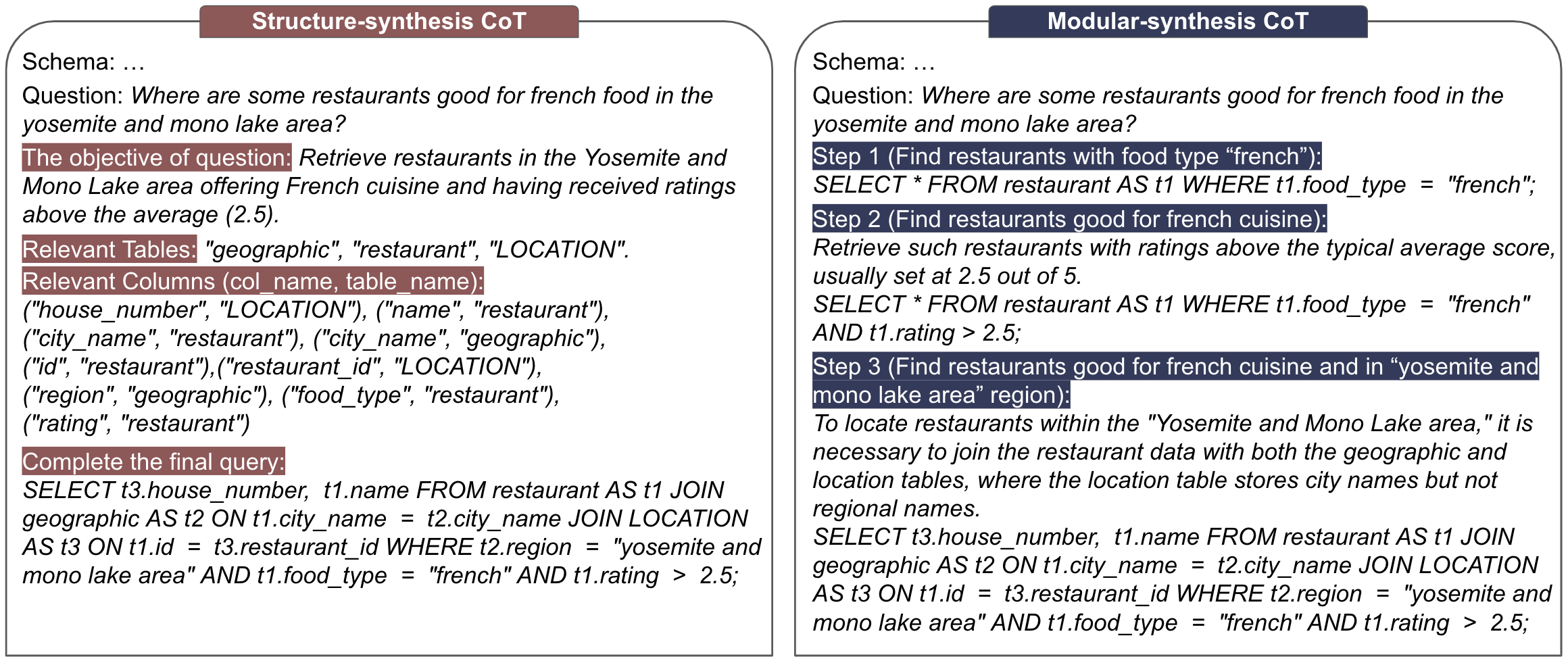}
  \caption{Examples of structure-synthesis (SS) and modular-synthesis (MS) CoT prompting.}
  \label{figure: cot-examples}
\end{figure}

Chain-of-thought (CoT; \cite{wei2022chain})is a prompting technique that guides an LLM to generate intermediate steps or explanations when solving complex reasoning problems. This approach enhances the model's interpretability and often leads to more accurate and reliable outcomes. Here, we introduce two different automated CoT prompting methods to decompose the original complex question into a multi-steps thinking process: structure-synthesis (SS) and modular-synthesis (MS) CoT. Each exemplar in the few-shot prompting $\sigma(\cdot,\cdot, \cdot)$ is augmented with a thought process $q' \in \textit{CoT}(\mathit{Q'})$ for the associated answer, as shown below, thereby improving the model's performance. Examples of these two CoT methods are illustrated in Figure \ref{figure: cot-examples}.

\begin{equation}\label{equation:cot}
\begin{aligned}
\max\limits_{y, \sigma, \mathit{Q'}, \textit{CoT}} & \mathbb{P}_{\mathcal{M}}(y|\sigma(q, D, \textit{CoT}(\mathit{Q'}))), 
\\
& \text{s.t.} \quad |\mathit{Q'}| = k, \mathit{Q'} \subset \mathit{Q},\textit{CoT} \in \{\textit{SS}, \textit{MS}\}.
\end{aligned}
\end{equation}

\paragraph{Structure-synthesis CoT} 

The SS CoT prompting decomposes the user's question into a structured sequence of data retrieval steps. The step-by-step solution begins with a deep understanding of the primary objective, informed by the schema of the provided database. We design a zero-shot prompting module to instruct GPT-3.5-Turbo to both paraphrase and, if necessary, expand the question in alignment with the schema of the target database. This step goes beyond the simple textual surface of the question to uncover deeper, contextual meanings essential for formulating an accurate SQL query. The complete prompt is provided in Appendix \ref{appendix-SS}.  

Subsequently, SS conducts a matching operation to identify the tables and columns, which are essential SQL components for composing the correct query. In the final step, we demonstrate how to use these components to compose the correct query and conclude the example.

\paragraph{Modular-synthesis CoT}

MS CoT, on the other hand, segments the original complex question into simpler, more manageable sub-questions, or modules, that incrementally build towards the final query. Each module addresses a specific aspect of the problem — for example, as illustrated in Figure \ref{figure: cot-examples}, the process begins by identifying restaurants based on the type of cuisine, filtering by rating, and concludes by refining the search to include only those within a specific geographic area. This step-by-step modular decomposition is generated by zero-shot prompting of GPT-3.5-Turbo, which is iteratively executed within a while loop as outlined in Algorithm \ref{algorithm: mscot}. Detailed information about the zero-shot \emph{firstSubQuestion} and \emph{nextStep} prompting is available in Appendix \ref{appendix-MS}.

\begin{algorithm}
\caption{Modular-synthesis CoT Prompting}
\begin{algorithmic}[1]
\State \textbf{Input:} Schema $\mathcal{S}$, Question $q$, Maximum Epochs $E$
\State \textbf{Output:} List of all subquestions $L_{subQ}$ and corresponding SQL queries $L_{sql}$
\State Initialize $L_{subQ}$ and $L_{sql}$ as empty lists, epoch $e$ to 1, $isValid$ to \textbf{false}
\State $q_{sub}, q_{sql} \gets$ firstSubQuestion prompt with $(\mathcal{S}, q)$
\State Append $q_{sub}$ to $L_{subQ}$ and $q_{sql}$ to $L_{sql}$
\State $isValid \gets$ Check if $q_{sql}$ correctly answers $q$ 

\While{not $isValid$ and $e < E$}
    \State $q_{sub}, q_{sql} \gets$ nextStep prompt with $(\mathcal{S}, q, q_{sub}, q_{sql})$ 
    \State Append $q_{sub}$ to $L_{subQ}$ and $q_{sql}$ to $L_{sql}$
    \State $isValid \gets$ Check if $L_{sql}$ solves $q$
    \State $e \gets$ $e + 1$
\EndWhile
\end{algorithmic}
\label{algorithm: mscot}
\end{algorithm}

\subsection{Self-correction}

\begin{figure}[ht]
  \centering
  \includegraphics[width=\textwidth]{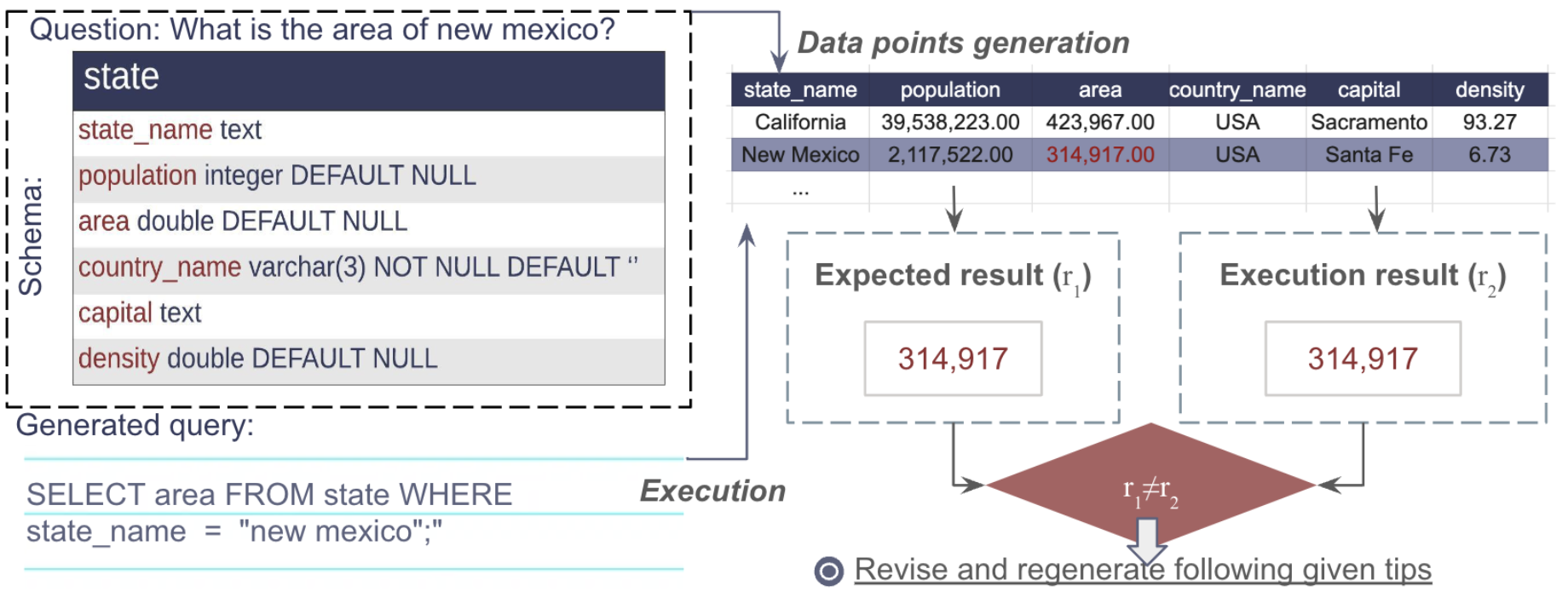}
  \caption{Illustration of the Self-Correction Component.}
  \label{fig: self-correction}
\end{figure}

In the same way that SQL experts review and revise their queries to guarantee precision, our method includes a self-correcting component, denoted as $SC$, that autonomously evaluates and refines its SQL outputs. While several intuitive strategies exist for implementing self-correction, our selected method is designed to execute the generated query on a crafted small-scale database, as illustrated in Figure \ref{fig: self-correction}. This database is purposefully constructed to contain some expected results given the input natural language questions. If any discrepancies, such as missing data, incorrect data retrieval, or syntax errors, are detected between the query output and the expected results, the component automatically flags the query for further revision.

Recognizing the potential prohibition in accessing real data for testing, we generate \emph{synthetic} yet plausible data points that are both manageable and reflective of the actual databases. Initially, we instruct GPT-3.5-Turbo to familiarize itself with the database schema and the input question and then generate 5 example data points (in SQL query format) for each table in the database.  To ensure the effectiveness of the execution validation process and to prevent scenarios where the expected results are null — which could obscure whether the model has produced a functional query — we prompt GPT-3.5-Turbo to demonstrate the expected outcomes from the synthetic database composed of the generated data points. The tested query can then be executed upon these synthetic data points, and the output can be compared with the expected outcomes to check for consistency.

In the revision step, we prompt GPT-3.5-Turbo to refine its generated query by supplementing the original input (the database schema and question) not only with the previously generated query that failed during execution testing but also with a set of human-crafted tips. Eq. (\ref{equation:ori}) turns to:

\begin{equation}
\begin{aligned}
\max\limits_{y, \sigma, \mathit{Q'}} &\mathbb{P}_{\mathcal{M}}(\mathcal{R}_{tips}(y)|\sigma(q, D, \mathit{Q'})), \\
& s.t.\quad |\mathit{Q'}| = k,\mathit{Q'} \subset \mathit{Q}
\end{aligned}
\end{equation}
Here, $\mathcal{R}_{tips}(y)$ represents the refinement of the generated query based on the provided tips. These tips aim to correct common errors observed in the model's preliminary outputs from a few training examples, ensuring that the evaluation data remains untouched. The recommended corrections include eliminating redundant joins, avoiding unnecessary nested queries, and including essential table joins. Additionally, we advise replacing the expression `<>' with `!=', and for questions targeting extreme values, such as the minimum, maximum, youngest, or oldest, using `LIMIT' in conjunction with `ORDER BY' or directly employing the `MIN' or `MAX' functions. Detailed prompting for data point generation and the revision process, including the tip details, are documented in Appendix \ref{appendix-SC}.

\subsection{Ensemble refinement}

Traditional ensemble machine learning models improve overall predictive performance by aggregating multiple weak predictors. Similarly, we adopt this strategy by using diverse prompts to generate multiple SQL queries for a given question, and request GPT-3.5-Turbo to identify the most accurate one. If none of the provided queries are correct, GPT is prompted to generate a new one. The rationale behind the approach is straightforward. Different prompts can induce various logical reasoning paths that might arrive at the same correct answer. By exposing GPT to the inherent variability of these potential solutions, we enable the model not only to evaluate the accuracy of each solution but also to engage in creative problem-solving, thereby improving the accuracy of the query generation process and enhancing the model's ability to handle complex query scenarios innovatively. Note that the provided queries do not necessarily need to be derived from CoT prompting or other few-shot prompting scenarios. In practical experiments, we have observed that incorporating queries generated from zero-shot prompting scenarios can significantly enhance the model's performance. The impact of varying the ensemble settings on performance is visualized in Appendix \ref{appendix-ensemble}.

\section{Experiments}
\label{experiments}

\paragraph{Datasets}
We conduct our experiments on the Spider\cite{yu2018spider} development (\textit{dev}) set. Spider is a large-scale, cross-domain Text-to-SQL benchmark designed to evaluate natural language interfaces across various databases. The dataset features instances where each instance includes a natural language question on a specific database, along with the corresponding SQL query. The training split has 8,659 instances; the development split has 1,034. All instances, including those in the test set, are spread across 200 databases without any overlap between the databases in different splits. The queries vary in complexity, from simple, straightforward ones to those requiring nested queries for retrieval, and are categorized into four levels of difficulty: \textit{easy, medium, hard,} and \textit{extra (hard)}. Specifically, in the \textit{dev} set, there are 248 queries classified as \textit{easy}, 446 as \textit{medium}, 174 as \textit{hard}, and 166 as \textit{extra hard}.
\paragraph{Metrics}
We follow the established Spider evaluation protocol \footnote{\url{https://github.com/taoyds/spider/blob/master/evaluation.py}} and other prior work by using \emph{execution accuracy} (EA) and component matching accuracy metrics in the evaluation. The EA metric compares the execution output of a query to the expected result of the ground truth query, providing a precise evaluation of model performance given that multiple SQL expressions can yield the same satisfactory results for a question. The matching metric, on the other hand, directly compares the generated query to the ground truth query partially, for each SQL clause such as SELECT and WHERE, measuring the exact token-level match between them. This analysis offers insights into the differences in expression preferences between human developers and the evaluated models. In \textit{Results}, we primarily focus on EA, as it is a more objective metric of success, while matching is utilized for error analysis.

\paragraph{Prompting}
In the experiments, we use GPT-3.5-Turbo and adopt the code representation (CR)\cite{chang2023prompt, nan2023enhancing, gao2023text} for schema and question representation in the Text-to-SQL task. This approach directly presents ``CREAT TABLE'' SQL statements to ChatGPT, enabling a thorough understanding of the database schema. Natural language questions are integrated as comments within the prompts, positioned beneath the schema. CR distinguishes itself from other representations by providing details essential for database construction, such as column types and primary/foreign keys, which enhance the comprehension needed to generate correct SQL queries.

For \emph{example selection} in the in-context learning setting, we use several methods. \textit{Question Similarity Selection }($QTS_S$)\cite{liu2022makes} selects $k$ examples based on similarities in question sentence embeddings. We developed a new encoder-based technique called \textit{Full-information Selection} ($FIS_S$), which analyzes embeddings of the database schema, masked question, and masked query, where domain-specific information is obscured. This approach provides a comprehensive method for selecting examples by taking into account all relevant textual elements.

Additionally, rather than relying on embedding similarity for choosing few-shot examples, we implement a novel strategy $Auto_S$, which prompts GPT-3.5-Turbo to identify the set of examples that are most relevant to the target question. Examples are categorized by difficulty level, and a diverse collection is randomly selected from various datasets. These selected examples are then used as input prompts for GPT, enabling it to choose those that are optimally aligned with the specific question.

For the ensemble settings, we use queries generated under four different prompting methods. The first is generated using zero-shot prompting. The next two responses are produced using SS CoT prompting with the $Auto_S$ and $FIS_S$ example selection approaches, respectively. The final response is generated using MS CoT prompting with the $Auto_S$ example selection approach. We also explored other combinations as ensemble configurations in the experiments, with the results detailed in Appendix \ref{appendix-ensemble}. Note that this study primarily focuses on GPT-3.5-Turbo and the Spider leaderboard, without evaluating the SelECT-SQL using more advanced models or other benchmarks. We plan to explore SelECT-SQL with more advanced LLMs on additional benchmarks in the future.

\paragraph{Baselines}
We compare our method against baselines listed on the official Spider website, all of which were evaluated on the same \textit{dev} set:
\begin{itemize}
    \item ChatGPT-SQL: Liu et al.\cite{liu2023comprehensive} examined GPT-3.5-Turbo's zero-shot Text-to-SQL capabilities by providing only the Sqlite SQL tables and their properties and then asked GPT to generate complete Sqlite SQL queries.
    \item RAT-SQL: Wang et al.\cite{wang2020rat} introduced a unified framework based on the BERT encoder, using relation-aware self-attention to improve schema encoding, linking, and representation. 
    \item PICARD: Scholak et al.\cite{scholak2021picard} proposed a method that enhances PLM decoders, such as T5, by using incremental parsing to constrain auto-regressive decoding, effectively selecting plausible tokens at each step.
    \item Graphix: Li et al.\cite{li2023graphix} proposed GRAPHIX-T5, a new architecture that augments the pre-trained T5 model by integrating structural inductive biases into Text-to-SQL parsers, improving the model's multi-hop reasoning abilities.
    \item SC-prompt: Gu et al.\cite{gu2023few} divided Text-to-SQL translation into two phases: structure and content. In the structure phase, a PLM is tasked with generating the framework of an SQL query. The content phase then directs a PLM to fill this framework with specific values, completing the query.
    \item C3: Dong et al.\cite{dong2023c3} provided a Text-to-SQL method that integrates clear prompting, calibration with hints, and consistent output checking, each designed to refine inputs, mitigate model biases, and enhance outputs, respectively.
    \item DAIL-SQL: Gao et al.\cite{gao2023text} developed a new Text-to-SQL prompt engineering technique DAIL-SQL, identified effective question representations and key components in prompts to utilize the in-context learning abilities of LLMs for the Text-to-SQL task.
\end{itemize}

\section{Results}\label{results}

\begin{figure}[ht]
  \centering
  \includegraphics[width=\textwidth]{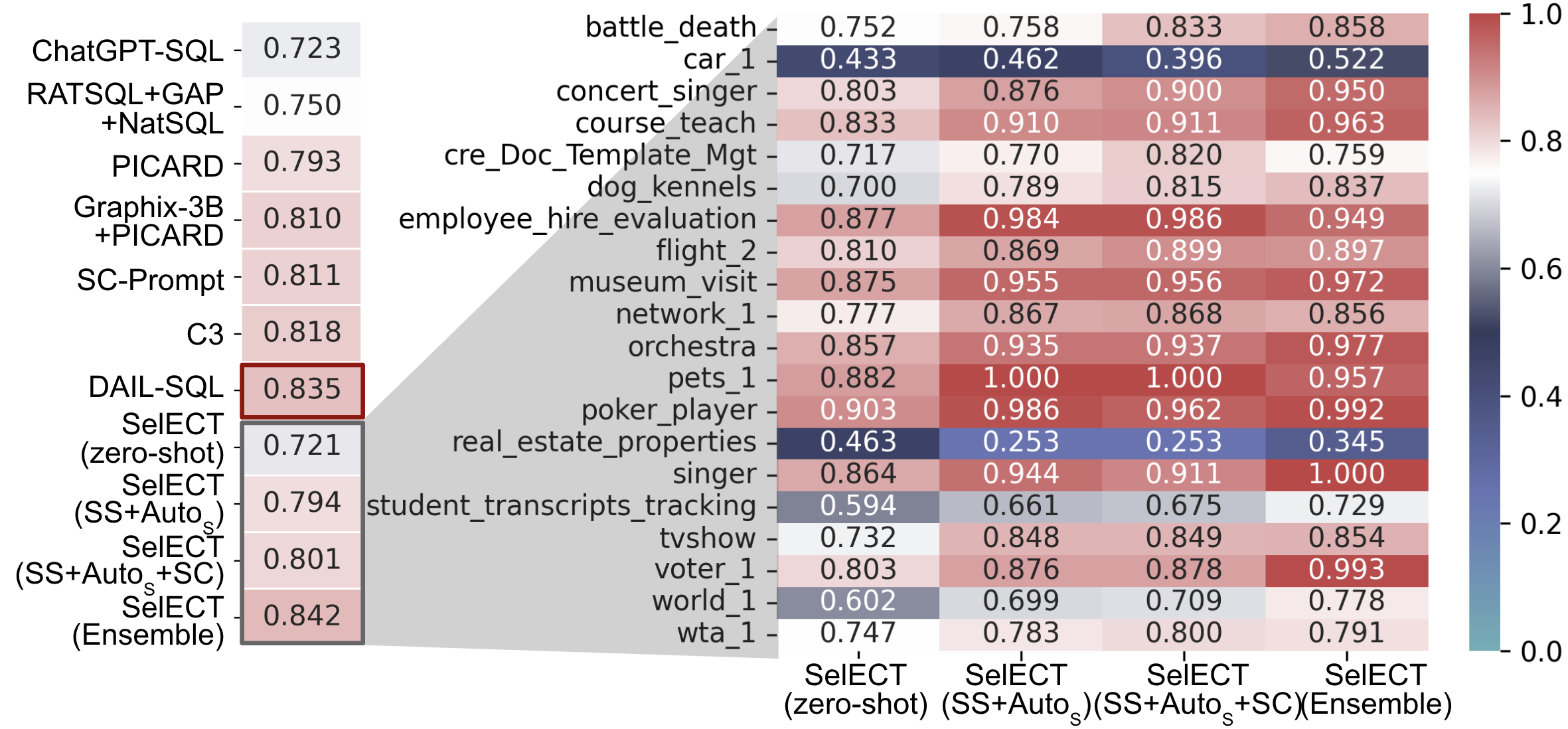}
  \caption{Execution accuracy of baselines and  GPT-3.5-Turbo implemented with SelECT-SQL on Spider \textit{dev} set. The DAIL-SQL baseline method, highlighted with a red border, employs the GPT-4 engine, while other baselines utilize zero-shot or few-shot prompted GPT-3.5-Turbo or fine-tuned T5. The left side shows the overall accuracy of both our methods and the baselines, while the right side presents the detailed performance of SelECT-SQL across different databases in Spider-dev.}
  \label{fig: execution_accuracy}
\end{figure}

\paragraph{Overall execution accuracy}
We report the performance of our method and baseline methods on the Spider leaderboard in Figure \ref{fig: execution_accuracy}.  Our method outperforms those that use fine-tuned T5-3B and BERT models, as well as GPT-3.5-Turbo and GPT-4 models employing zero-shot or in-context learning settings. Specifically, compared to the C3 method, which employs a zero-shot GPT-3.5-Turbo with a self-consistency component (including 20 generations), our approach uses approximately 80\% fewer tokens; compared to DAIL-SQL, which uses GPT-4 engines with few-shot learning, our method reduces costs by 20x times, making it more cost-effective.

\paragraph{Ablation Study} Table \ref{tab:ablation_study} presents the ablation study on various components of SelECT-SQL, revealing that each component significantly contributes to the overall performance. Removing the self-correction (SC) or ensemble components results in an average performance decline of 2.7\% and 1.2\%, respectively, across different few-shot settings ($k={1, 3}$). Incorporating the one-shot SelECT-SQL ensemble component increases GPT-3.5-Turbo's execution accuracy by 4.1\%. Additionally, the study indicates that the basic few-shot settings are crucial; three-shot scenarios typically decrease performance by 2.2\% compared to one-shot scenarios. This highlights the importance of each component's specific configuration in enhancing the model's accuracy and efficiency.

\begin{table}[ht]

    \centering
    \caption{Ablation studies and in-context learning performance analysis of GPT-3.5-Turbo implementing SelECT-SQL on the Spider Dev Set. The table reports the comparison between one-shot and three-shot settings, where the first value represents the one-shot prompting EA result, and applying the value in brackets to this result yields the three-shot setting EA. Positive values in the brackets, indicating higher performance in the three-shot setting compared to the one-shot setting, are italicized. The performance under different in-context learning settings is significantly different from zero-shot performance at the 99\% confidence level.}
    \begin{subtable}{\linewidth}
        \centering

\caption{Ablation studies of self-correction (SC) and ensemble techniques. The SelECT-SQL line shows the performance of ensemble query generation using tips-augmented self-corrected responses from the four prompting methods discussed in the Experiments section. We report the performance of self-corrected responses generated under SS CoT prompting with the $Auto_S$ example selection approach, the self-corrected $SS$ + $Auto_S$ (as detailed in \ref{tab:example_selection}), in the w/o Ensemble line. The w/o SC line presents the ensemble results of the four prompting methods without self-correction.} \footnotesize
\begin{tabular}{p{1.83cm}p{1.97cm}p{1.97cm}p{1.95cm}p{1.95cm}p{2.0cm}}
\hline
             & Easy           & Medium         & Hard          & Extra          & Overall        \\ \hline
SelECT-SQL   & \textbf{0.931 (-0.012)} & \textbf{0.886 (-0.045)} & \textbf{0.810 (-0.025)} & \textbf{0.627 (-0.037)} & \textbf{0.842 (-0.032)} \\ \hdashline
w/o Ensemble & 0.903 (-0.040)          & 0.850 (-0.011)          & 0.747 (-0.046)          & 0.572 (-0.030)          & 0.801 (-0.027)          \\ \hdashline
w/o SC       & 0.923 (-0.004)          & 0.863 (+\textit{0.000})    & 0.799 (+\textit{0.011})   & 0.627 (-0.042)          & 0.829 (-0.006)          \\ \hline
\end{tabular}
        
        \label{tab:ablation_study}
    \end{subtable}
    \hfill
    \begin{subtable}{\linewidth}
        \centering
        \caption{Effects of different in-context learning settings, specifically various example selection and CoT methods. }
        \centering \footnotesize
\begin{tabular}{p{1.83cm}p{1.97cm}p{1.97cm}p{1.95cm}p{1.95cm}p{2.0cm}}
\hline
             & Easy                    & Medium                  & Hard                    & Extra                   & Overall                 \\ \hline
zero-shot    & 0.875                   & 0.713                   & 0.724                   & 0.512                   & 0.721                   \\ \hline
$QTS_S$       & 0.863 (+\textit{0.004}) & 0.726 (-0.004)          & 0.690 (+\textit{0.006}) & 0.494 (+\textit{0.024}) & 0.716 (+\textit{0.004}) \\\hdashline
$FIS_S$       & 0.819 (+\textit{0.056}) & 0.771 (-0.078)          & 0.684 (+\textit{0.011}) & 0.524 (-0.030)          & 0.728 (-0.023)          \\\hdashline
$Auto_S$      & 0.827 (+\textit{0.024}) & 0.767 (-0.027)          & 0.701 (-0.109)          & 0.524 (-0.030)          & 0.731 (-0.029)          \\\hline
$MS$ + $Auto_S$ & 0.823 (-0.004)          & 0.785 (-0.049)          & 0.695 (-0.195)          & 0.530 (-0.066)          & 0.738 (-0.066)          \\\hdashline
$SS$ + $QTS_S$  & 0.847 (+\textit{0.004}) & 0.823 (+\textit{0.000}) & 0.701 (+\textit{0.000}) & 0.536 (+\textit{0.006}) & 0.762 (+\textit{0.002}) \\\hdashline
$SS$ + $FIS_S$  & 0.847 (+\textit{0.012}) & 0.814 (-0.020)          & \textbf{0.753} (-0.046)          & 0.548 (-0.018)          & 0.769 (-0.016)          \\\hdashline
$SS$ + $Auto_S$ & \textbf{0.907} (-0.020)          & \textbf{0.836} (-0.024)          & 0.741 (-0.063)          & \textbf{0.566} (-0.012)          & \textbf{0.794} (-0.028)          \\\hline
\end{tabular}
        
        \label{tab:example_selection}
    \end{subtable}
    
    \label{tab:component_analysis}
\end{table}

In Table \ref{tab:example_selection}, we further investigate the effects of various in-context learning settings on the performance of the GPT-3.5-Turbo engine in Text-to-SQL tasks. Our findings indicate that, still, the GPT-3.5-Turbo engine with one-shot settings consistently outperforms the three-shot settings, except when using the question similarity selection ($QTS_S$) method. One-shot scenarios exhibit an average increase of 2.2\% in execution accuracy compared to three-shot scenarios. Under one-shot settings, the proposed example selection method, $Auto_S$ shows enhanced performance compared to other example selection methods, increasing the GPT-3.5-Turbo's execution accuracy by 1\%. This improvement is particularly notable in medium-level questions (5.4\%) and extra-hard-level questions (1.2\%) on the Spider dev set, compared to zero-shot setting.

CoT prompting is observed to significantly boost the performance of GPT-3.5-Turbo. The one-shot MS and SS CoT methods increase execution accuracy by 1.7\% and 5.4\%, respectively. Given SS CoT's structured step-by-step solutions, which facilitate the automatic generation of solution steps as examples, it proves to be an efficient and cost-effective prompting method for Text-to-SQL tasks using GPT engines. The $Auto_S$ method, when combined with CoT prompting, outperforms other methods, enhancing performance by 7.3\% compared to zero-shot scenarios. 

We also experimented with zero-shot CoT methods. The classical approach\cite{kojima2022large}, prompting with ``let's think step by step,'' achieved an execution accuracy of 73.9\%. Additionally, a more instructive approach, requiring the model to consider the objective, required tables, and columns before generating the response query, resulted in an accuracy of 72.1\%. These results suggest that while directing the model to outline thinking steps can help, overly prescriptive prompts without allowing for open-ended analysis may reduce performance compared to more open-ended analytical prompts.


\section{Error analysis}\label{error_analysis}

The execution accuracy of GPT-Turbo-3.5 with SelECT-SQL prompting varies across different databases, as shown on the right-hand side of Figure \ref{fig: execution_accuracy}. It achieves an accuracy above 0.9 on 9 out of 20 databases in the Spider dev set, particularly in domains such as singers, poker players, pets, orchestra, and employee evaluations. In some instances, the engine answered all related questions correctly. However, for car and real estate property databases, which include more questions at the hard or extra hard levels, the execution accuracy drops to 0.5 or below. The average number of tables in car and real estate databases is 5.5, whereas it is only 3 in the first group (singers, poker players, etc.). Evidence suggests that the relatively complex schema structures in certain domains may impede the further improvement of GPT engines.

\begin{figure}[ht]
  \centering
  \includegraphics[width=0.9\textwidth]{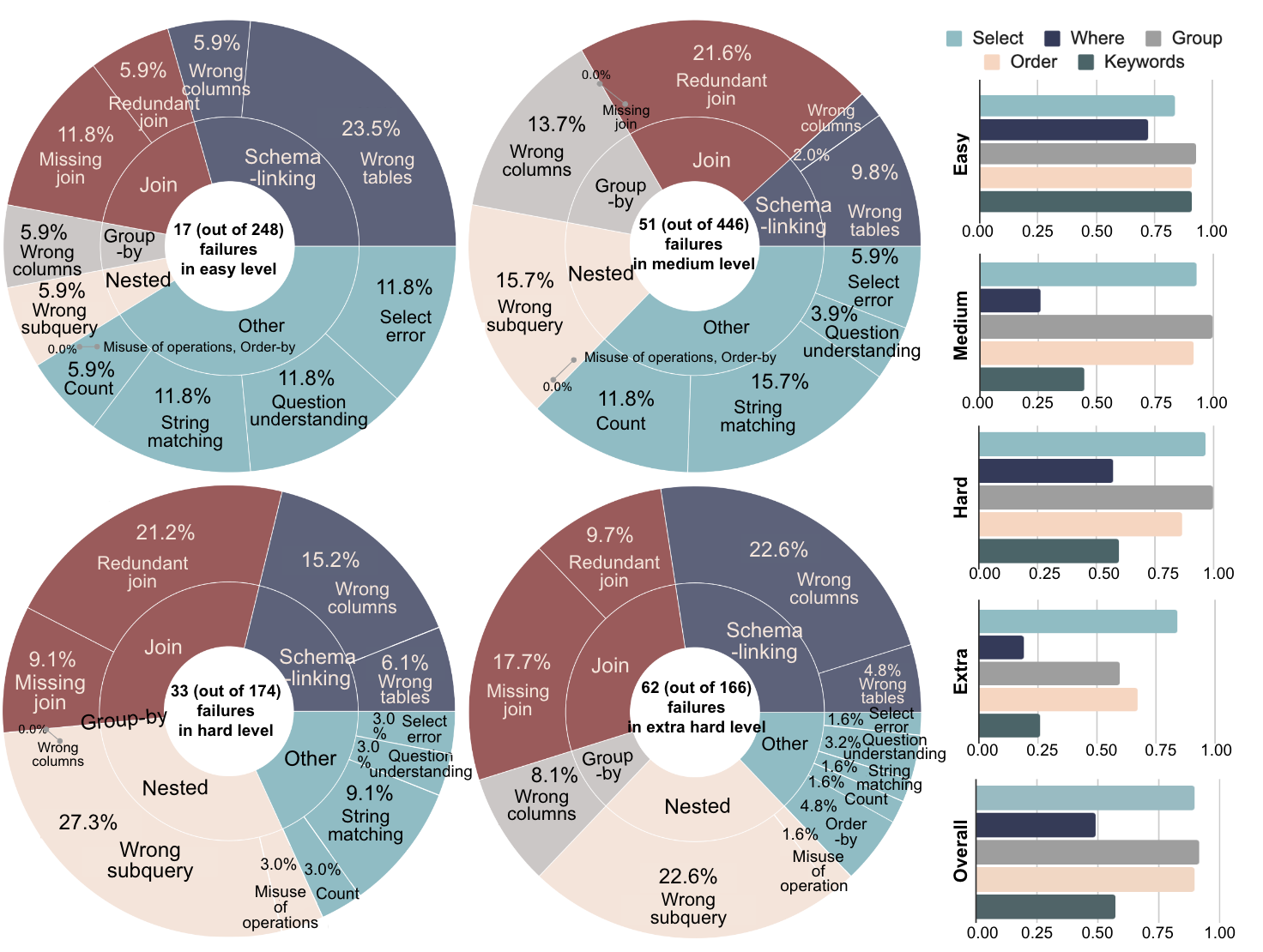}
  \caption{Error Distribution in ChatGPT-Generated Queries on the Spider-dev set. We report the partial clause component matching accuracy on the left side as a reference. Note that the partial accuracy is loosely correlated with the error portion shown in the distribution pie chart, as the missing matching of partial clauses may not result in an incorrect query. There may exist several correct queries that can solve the same question.}
  \label{fig: erroranalysis}
\end{figure}

To gain clearer insights into the limitations of GPT engines and pinpoint failure areas, we conducted an error analysis, visualized in Figure \ref{fig: erroranalysis}. According to the literature \cite{pourreza2023din}, errors are categorized into six types: schema-linking, join, group-by, nested, invalid, and other. Our analysis reveals that, with SelECT-SQL, GPT-3.5-Turbo no longer suffers from \emph{invalid syntax} errors. To provide a more intuitive understanding of GPT-3.5-Turbo's failures, we refined the error categories, especially within the \emph{join} and \emph{other} groups, reducing the number of error types from six to five.

In the \emph{join} category, we focused on whether the engine missed join operations or included unnecessary joins. Errors involving incorrect joins based on wrong table and column names were reclassified as schema-linking errors, as they stem from incorrect schema understanding. In the \emph{other} category, we include cases that do not fit under any of the other specific categories but specifically focus on whether the generated queries accurately conveyed the information requested by the original question. This includes verifying if the engine correctly selected and counted the required columns, comprehended the true meaning of the question, and effectively handled string-matching syntax to produce the expected output. Detailed definitions of the error categories can be found in Appendix \ref{appendix-error}.

In Figure \ref{fig: erroranalysis}, it is evident that as the complexity of questions increases, the frequency of errors caused by inaccurate join operations rises. For extra hard questions, errors typically arise from missing essential joins between certain tables, constituting 17.7\% of the errors, which results in insufficiently narrowed query results and consequently incorrect outcomes. Nested-related errors also become more common in harder questions, often due to the misuse of set operations and the generation of incorrect subqueries within nested structures. Despite our self-correction component's prompting tips, which include careful selection of tables and columns in join operations and revising subqueries in nested queries, 56.1\% of errors in hard and extra hard questions still stem from incorrect nested and join operations. The manual review also reveals that, compared to human-crafted ground truth, GPT engines show a clear preference for nested (incorrect) operations over joins, even though joins are generally more effective in database analysis.

Notably, string matching and select-related errors frequently occur in easy and medium questions. These errors typically involve the erroneous use of string matching patterns in queries or the selection of inappropriate columns, often resulting in the presentation of more information than necessary. For instance, when asked, ``What is the number of cars with a horsepower greater than 150?'' the ground truth query is `SELECT count(*) FROM CARS\_DATA WHERE horsepower > 150;', whereas the GPT engine generates `SELECT COUNT(*) FROM CARS\_DATA WHERE CAST(REPLACE(Horsepower, ' hp', '') AS INTEGER) > 150;', illustrating a string matching issue. When asked ``Describe the section h.'', the ground truth query is `SELECT section\_description FROM Sections WHERE section\_name = 'h';', while the generated query is `SELECT * FROM Sections WHERE UPPER(section\_name) = UPPER('h');', exemplifying a selection issue. The partial component matching accuracy shown on the left side of Figure \ref{fig: erroranalysis}, from another aspect, verifies that mismatches between WHERE clauses and keywords components in the ground truth query and the generated query are frequent in easy- and medium-level questions. These errors could be attributed to the model's lack of clarity on the detailed data points based on the schema, which might be mitigated by presenting few example data points to the reasoning engine.

In some cases, errors arise from misunderstanding the input question or lacking inherent commonsense knowledge. For instance, when asked, ``What is the best rank of losers across all matches?'' humans naturally interpret ``best rank'' as the minimum ranking number due to ascending order ranking, resulting in the query `SELECT MIN(loser\_rank) FROM matches;'. Conversely, GPT engines interpret the question more intuitively, leading to the incorrect query `SELECT MAX(loser\_rank) FROM matches;'. Fine-tuning with commonsense knowledge or designing specific prompts to induce deeper reasoning may further enhance performance. We leave the exploration of plausible solutions as an open question for further research.

\section{Conclusion}\label{conclusion}

In this paper, we introduced SelECT-SQL, a novel self-correcting ensemble CoT prompting method for Text-to-SQL. By generating ensemble query responses in a tips-augmented self-correction setting and using zero-shot prompted and one-shot SS CoT examples, we achieved 84.2\% execution accuracy on the Spider-dev leaderboard.  This demonstrates the effectiveness of SelECT-SQL in employing CoT prompting and self-correction, showing substantial performance gains with token-efficient one-shot learning. SelECT-SQL enhances GPT-3.5-Turbo's capabilities, delivering results comparable to or surpassing state-of-the-art GPT-4 methods.

\newpage
\bibliographystyle{unsrt}  
\bibliography{references}  

\begin{thebibliography}{10}

\bibitem{deng2022recent}
Naihao Deng, Yulong Chen, and Yue Zhang.
\newblock Recent advances in text-to-sql: A survey of what we have and what we
  expect.
\newblock In {\em Proceedings of the 29th International Conference on
  Computational Linguistics}, pages 2166--2187, 2022.

\bibitem{katsogiannis2023survey}
George Katsogiannis-Meimarakis and Georgia Koutrika.
\newblock A survey on deep learning approaches for text-to-sql.
\newblock {\em The VLDB Journal}, 32(4):905--936, 2023.

\bibitem{popescu2003towards}
Ana-Maria Popescu, Oren Etzioni, and Henry Kautz.
\newblock Towards a theory of natural language interfaces to databases.
\newblock In {\em Proceedings of the 8th international conference on
  Intelligent user interfaces}, pages 149--157, 2003.

\bibitem{popescu2004modern}
Ana-Maria Popescu, Alex Armanasu, Oren Etzioni, David Ko, and Alexander Yates.
\newblock Modern natural language interfaces to databases: Composing
  statistical parsing with semantic tractability.
\newblock In {\em COLING 2004: Proceedings of the 20th International Conference
  on Computational Linguistics}, pages 141--147, 2004.

\bibitem{li2014constructing}
Fei Li and Hosagrahar~V Jagadish.
\newblock Constructing an interactive natural language interface for relational
  databases.
\newblock {\em Proceedings of the VLDB Endowment}, 8(1):73--84, 2014.

\bibitem{stratica2005using}
Niculae Stratica, Leila Kosseim, and Bipin~C Desai.
\newblock Using semantic templates for a natural language interface to the
  cindi virtual library.
\newblock {\em Data \& Knowledge Engineering}, 55(1):4--19, 2005.

\bibitem{zhong2017seq2sql}
Victor Zhong, Caiming Xiong, and Richard Socher.
\newblock Seq2sql: Generating structured queries from natural language using
  reinforcement learning.
\newblock {\em arXiv preprint arXiv:1709.00103}, 2017.

\bibitem{yu2018spider}
Tao Yu, Rui Zhang, Kai Yang, Michihiro Yasunaga, Dongxu Wang, Zifan Li, James
  Ma, Irene Li, Qingning Yao, Shanelle Roman, et~al.
\newblock Spider: A large-scale human-labeled dataset for complex and
  cross-domain semantic parsing and text-to-sql task.
\newblock In {\em Proceedings of the 2018 Conference on Empirical Methods in
  Natural Language Processing}, pages 3911--3921, 2018.

\bibitem{scholak2021picard}
Torsten Scholak, Nathan Schucher, and Dzmitry Bahdanau.
\newblock Picard: Parsing incrementally for constrained auto-regressive
  decoding from language models.
\newblock In {\em Proceedings of the 2021 Conference on Empirical Methods in
  Natural Language Processing}, pages 9895--9901, 2021.

\bibitem{xu2022sead}
Kuan Xu, Yongbo Wang, Yongliang Wang, Zihao Wang, Zujie Wen, and Yang Dong.
\newblock Sead: End-to-end text-to-sql generation with schema-aware denoising.
\newblock In {\em Findings of the Association for Computational Linguistics:
  NAACL 2022}, pages 1845--1853, 2022.

\bibitem{qi2022rasat}
Jiexing Qi, Jingyao Tang, Ziwei He, Xiangpeng Wan, Yu~Cheng, Chenghu Zhou,
  Xinbing Wang, Quanshi Zhang, and Zhouhan Lin.
\newblock Rasat: Integrating relational structures into pretrained seq2seq
  model for text-to-sql.
\newblock {\em arXiv preprint arXiv:2205.06983}, 2022.

\bibitem{li2023graphix}
Jinyang Li, Binyuan Hui, Reynold Cheng, Bowen Qin, Chenhao Ma, Nan Huo, Fei
  Huang, Wenyu Du, Luo Si, and Yongbin Li.
\newblock Graphix-t5: Mixing pre-trained transformers with graph-aware layers
  for text-to-sql parsing.
\newblock In {\em Proceedings of the AAAI Conference on Artificial
  Intelligence}, volume~37, pages 13076--13084, 2023.

\bibitem{gao2023retrieval}
Yunfan Gao, Yun Xiong, Xinyu Gao, Kangxiang Jia, Jinliu Pan, Yuxi Bi, Yi~Dai,
  Jiawei Sun, and Haofen Wang.
\newblock Retrieval-augmented generation for large language models: A survey.
\newblock {\em arXiv preprint arXiv:2312.10997}, 2023.

\bibitem{liu2023comprehensive}
Aiwei Liu, Xuming Hu, Lijie Wen, and Philip~S Yu.
\newblock A comprehensive evaluation of chatgpt's zero-shot text-to-sql
  capability.
\newblock 2023.

\bibitem{dong2023c3}
Xuemei Dong, Chao Zhang, Yuhang Ge, Yuren Mao, Yunjun Gao, Jinshu Lin, Dongfang
  Lou, et~al.
\newblock C3: Zero-shot text-to-sql with chatgpt.
\newblock {\em arXiv preprint arXiv:2307.07306}, 2023.

\bibitem{gu2023few}
Zihui Gu, Ju~Fan, Nan Tang, Lei Cao, Bowen Jia, Sam Madden, and Xiaoyong Du.
\newblock Few-shot text-to-sql translation using structure and content prompt
  learning.
\newblock 1(2):1--28, 2023.

\bibitem{gao2023text}
Dawei Gao, Haibin Wang, Yaliang Li, Xiuyu Sun, Yichen Qian, Bolin Ding, and
  Jingren Zhou.
\newblock Text-to-sql empowered by large language models: A benchmark
  evaluation.
\newblock 2023.

\bibitem{pourreza2023din}
Mohammadreza Pourreza and Davood Rafiei.
\newblock Din-sql: Decomposed in-context learning of text-to-sql with
  self-correction.
\newblock In {\em Thirty-seventh Conference on Neural Information Processing
  Systems}, 2023.

\bibitem{brown2020language}
Tom Brown, Benjamin Mann, Nick Ryder, Melanie Subbiah, Jared~D Kaplan, Prafulla
  Dhariwal, Arvind Neelakantan, Pranav Shyam, Girish Sastry, Amanda Askell,
  et~al.
\newblock Language models are few-shot learners.
\newblock {\em Advances in neural information processing systems},
  33:1877--1901, 2020.

\bibitem{wei2022chain}
Jason Wei, Xuezhi Wang, Dale Schuurmans, Maarten Bosma, Fei Xia, Ed~Chi, Quoc~V
  Le, Denny Zhou, et~al.
\newblock Chain-of-thought prompting elicits reasoning in large language
  models.
\newblock {\em Advances in neural information processing systems},
  35:24824--24837, 2022.

\bibitem{wang2022self}
Xuezhi Wang, Jason Wei, Dale Schuurmans, Quoc~V Le, Ed~H Chi, Sharan Narang,
  Aakanksha Chowdhery, and Denny Zhou.
\newblock Self-consistency improves chain of thought reasoning in language
  models.
\newblock In {\em The Eleventh International Conference on Learning
  Representations}, 2022.

\bibitem{li2023symbolic}
Liunian~Harold Li, Jack Hessel, Youngjae Yu, Xiang Ren, Kai-Wei Chang, and
  Yejin Choi.
\newblock Symbolic chain-of-thought distillation: Small models can also" think"
  step-by-step.
\newblock {\em arXiv preprint arXiv:2306.14050}, 2023.

\bibitem{zhao2023verify}
Ruochen Zhao, Xingxuan Li, Shafiq Joty, Chengwei Qin, and Lidong Bing.
\newblock Verify-and-edit: A knowledge-enhanced chain-of-thought framework.
\newblock {\em arXiv preprint arXiv:2305.03268}, 2023.

\bibitem{chang2023prompt}
Shuaichen Chang and Eric Fosler-Lussier.
\newblock How to prompt llms for text-to-sql: A study in zero-shot,
  single-domain, and cross-domain settings.
\newblock In {\em NeurIPS 2023 Second Table Representation Learning Workshop},
  2023.

\bibitem{nan2023enhancing}
Linyong Nan, Yilun Zhao, Weijin Zou, Narutatsu Ri, Jaesung Tae, Ellen Zhang,
  Arman Cohan, and Dragomir Radev.
\newblock Enhancing few-shot text-to-sql capabilities of large language models:
  A study on prompt design strategies.
\newblock 2023.

\bibitem{liu2022makes}
Jiachang Liu, Dinghan Shen, Yizhe Zhang, William~B Dolan, Lawrence Carin, and
  Weizhu Chen.
\newblock What makes good in-context examples for gpt-3?
\newblock In {\em Proceedings of Deep Learning Inside Out (DeeLIO 2022): The
  3rd Workshop on Knowledge Extraction and Integration for Deep Learning
  Architectures}, pages 100--114, 2022.

\bibitem{wang2020rat}
Bailin Wang, Richard Shin, Xiaodong Liu, Oleksandr Polozov, and Matthew
  Richardson.
\newblock Rat-sql: Relation-aware schema encoding and linking for text-to-sql
  parsers.
\newblock In {\em Proceedings of the 58th Annual Meeting of the Association for
  Computational Linguistics}, pages 7567--7578, 2020.

\bibitem{kojima2022large}
Takeshi Kojima, Shixiang~Shane Gu, Machel Reid, Yutaka Matsuo, and Yusuke
  Iwasawa.
\newblock Large language models are zero-shot reasoners.
\newblock {\em Advances in neural information processing systems},
  35:22199--22213, 2022.

\end{thebibliography}
\newpage

\section{Appendix / supplemental material}

\subsection{Compute resources}
The experiments were carried out using a MacBook Pro equipped with a 2.3 GHz 8-core i9 Core processor and 16GB of memory. The total cost of GPT-3.5-Turbo prompting for all experiments, inclusive of experiments not mentioned in the main text, amounted to approximately \$41.1. The longest duration required to prompt the spider dev set is 7176 seconds.

\subsection{Objective paraphrasing prompt for Structure-synthesis CoT prompting.}\label{appendix-SS}
\begin{figure}[h]
  \centering
  \includegraphics[width=\textwidth]{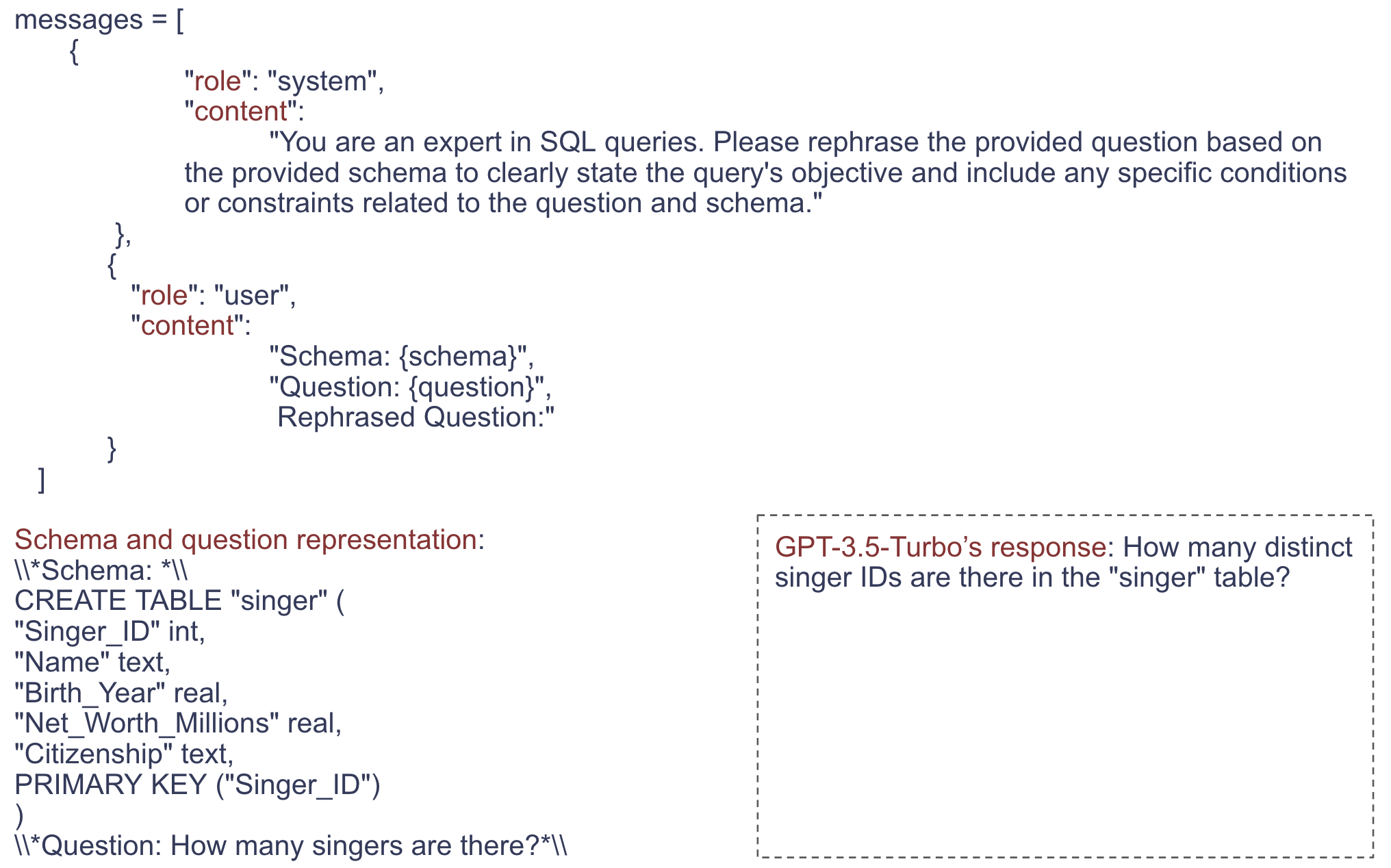}
  \caption{Objective paraphrasing prompt for SS CoT: Above, we provide the schema and question representation along with the detailed prompting message. An example response is shown on the right.}
  \label{figure: objPara}
\end{figure}

We present the zero-shot prompting for the paraphrasing objective used in SS CoT in Figure \ref{figure: objPara}.

\subsection{FirstSubQuestion and nextStep prmot for Modular-synthesis CoT prompting.}\label{appendix-MS}
\begin{figure}[h]
    \centering
    \begin{subfigure}[h]{\textwidth}
        \centering
        \includegraphics[width=\textwidth]{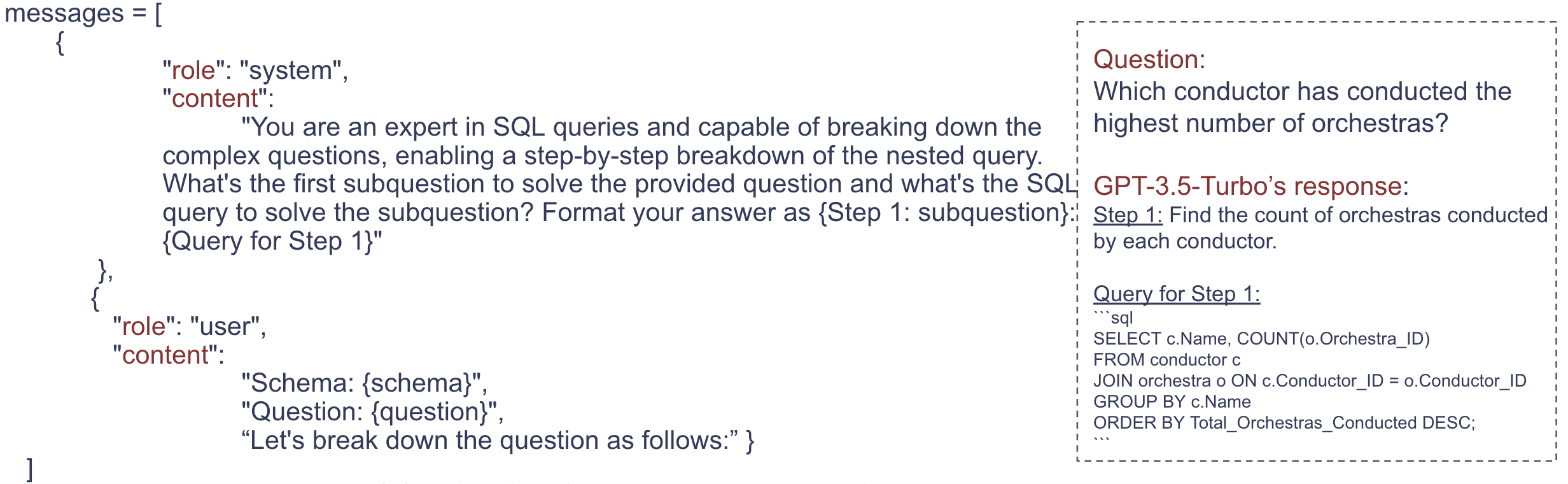}
        \caption{The firstSubQuestion prompt and GPT-3.5-Turbo’s response example.}
        \label{fig:firstsubq}
    \end{subfigure}
    \hfill
    \begin{subfigure}[h]{\textwidth}
        \centering
        \includegraphics[width=\textwidth]{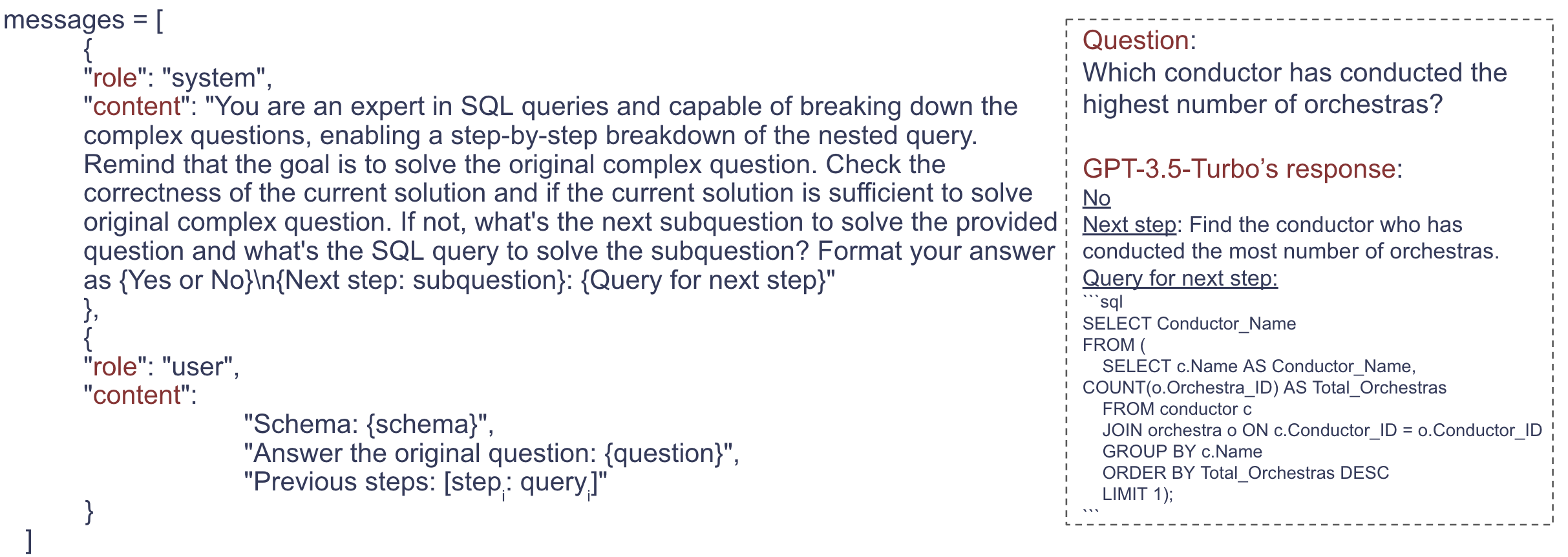}
        \caption{The nextStep prompt and GPT-3.5-Turbo’s response example.}
        \label{fig:nextstep}
    \end{subfigure}
    \caption{The firstSubQUestion and nextStep prompt for MS CoT. The example response is shown on the right.}
    \label{fig:MS-Cot}
\end{figure}

Figure \ref{fig:MS-Cot} presents the zero-shot prompting for the firstSubQuestion and nextStep decomposition utilized in MS CoT. In the experiments, we set the maximum epochs as 5.

\subsection{Data points generation prompting and human-crafted tips used in self-correction component.}\label{appendix-SC}

\begin{figure}[h]
    \centering
    \includegraphics[width=\textwidth]{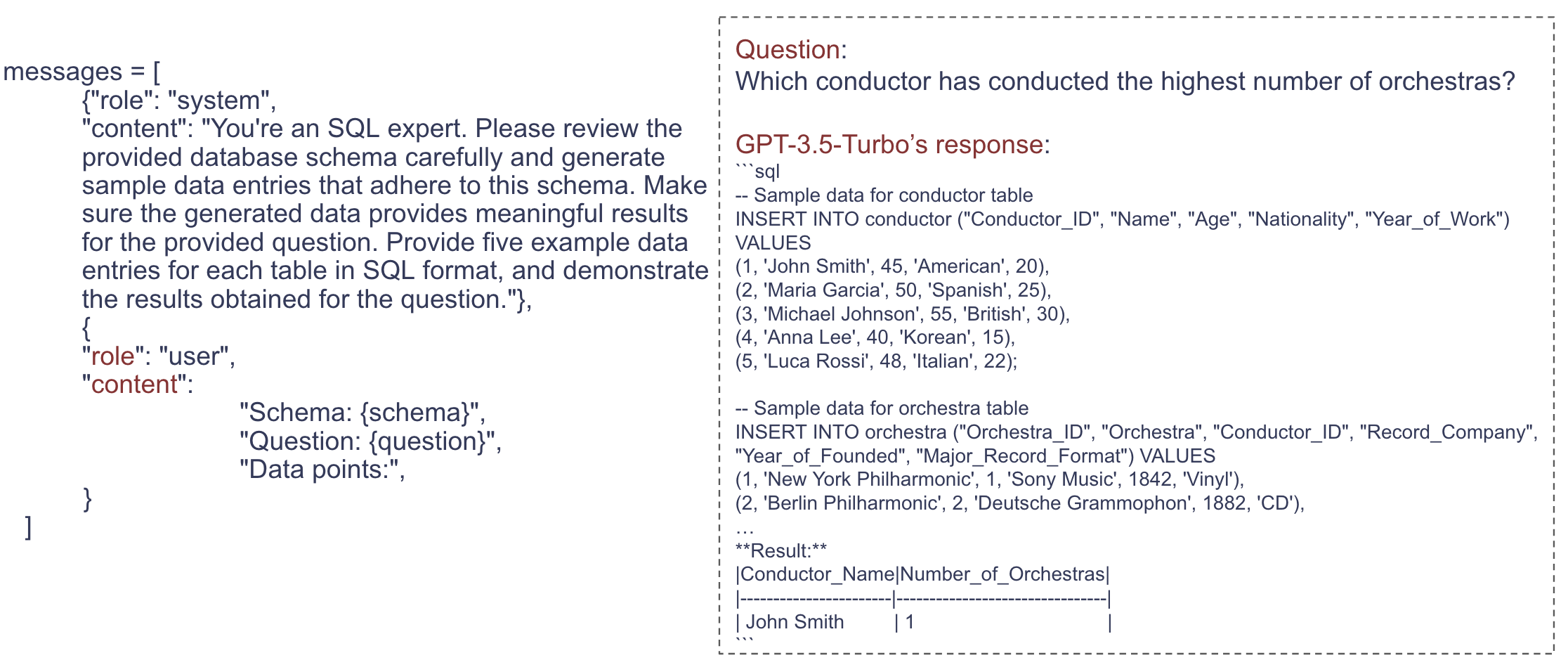}
    \caption{Data points generation prompting for self-correction component.}
    \label{fig:datageneration}
\end{figure}

Here are the tips used to revise and refine the generated query in the self-correction component:
\begin{itemize}
  \item When using the `GROUP BY' clause, consider prioritizing the primary key if it aligns logically with the requirements of question. Often, you may need to group by other columns to achieve meaningful data summaries and analyses.
  \item Use clauses `LEFT JOIN, `CAST', `REPLACE', `DATEDIFF', `IN', and `OR' judiciously, and replace `<>' with `!=' in your generation.
  \item Ensure that join operations are executed correctly by avoiding redundant joins, unnecessary nested queries, or missing essential joins between tables.
  \item Whenever possible, opt for join clauses instead of nested queries in your SQL statements.
  \item If applicable, prefer using COUNT(\*) over COUNT(column$\_$name) when you only need to count rows.
  \item For questions aimed at identifying extremes such as the youngest, oldest, or top minimum or maximum values, opt for employing LIMIT in conjunction with ORDER BY, or utilize the MIN or MAX functions directly. This avoids the complexity and performance issues associated with nested queries and is especially effective for managing large datasets efficiently.
  \item Please provide a case-insensitive SQL query by possibly incorporating the LOWER() or UPPER() functions within the SQL query.
\end{itemize}

\subsection{Ensemble settings.}\label{appendix-ensemble}
\begin{figure}[h]
    \centering
    \includegraphics[width=\textwidth]{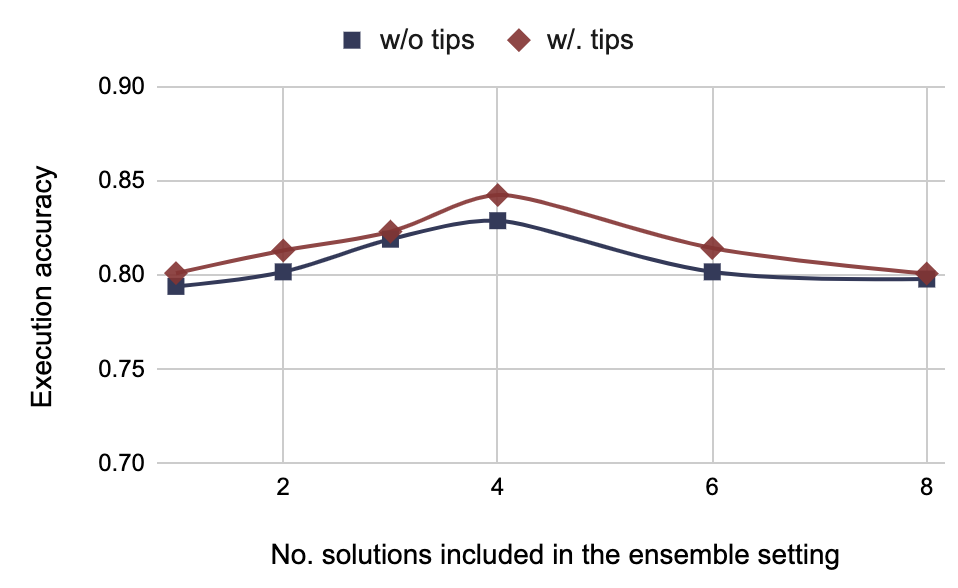}
    \caption{The overall execution accuracy performance of GPT-3.5-Turbo when exposed to varying numbers of potential solutions for selection.}
    \label{fig:ensemble}
\end{figure}

We report the execution accuracy of GPT-3.5-Turbo under different ensemble settings. The different ensemble settings refer to the number of potential solutions exposed to the engine for reference. We select results from the eight most diverse prompting strategies as potential solutions: zero-shot code representation, zero-shot code representation with schema explanation (where we ask GPT-3.5-Turbo to generate a detailed explanation of the schema, including what information is included in each table, what's the primary key and foreign key, etc.), one-shot (SS CoT+example selected by $FIS_S$), three-shot (SS CoT+example selected by $FIS_S$), one-shot (MS CoT+example selected by $Auto_S$), three-shot (MS CoT+$Auto_S$), one-shot (SS CoT+$Auto_S$), and three-shot (SS+$Auto_S$). The combinations of these prompting strategies under various settings of potential solutions are shown in Table \ref{table: ensemble_settings}. We present two versions of the ensemble results, self-corrected and non-self-corrected, visualized in Figure \ref{fig:ensemble}.

\begin{table}[]
\caption{The ensemble configuration.}
\begin{tabular}{|l|l|}
\hline
Ensemble setting (\# sols) & Prompting combination                                                                                                                                                                                                        \\ \hline
1                          & one-shot (SS CoT+$Auto_S$)                                                                                                                                                                                                   \\ \hline
2                          & one-shot (SS CoT+$Auto_S$;MS CoT+$Auto_S$)                                                                                                                                                                                   \\ \hline
3                          & \begin{tabular}[c]{@{}l@{}}one-shot ( SS CoT+$Auto_S$; MS CoT+$Auto_S$;SS CoT+$FIS_S$)\end{tabular}                                                                                                                            \\ \hline
4                          & \begin{tabular}[c]{@{}l@{}}zero-shot code representation;\\ one-shot (SS CoT+$Auto_S$; MS CoT+$Auto_S$;SS CoT+$FIS_S$)\end{tabular}                                                                                           \\ \hline
5                          & \begin{tabular}[c]{@{}l@{}}zero-shot ( code representation; w/. schema explanation)\\ one-shot ( SS CoT+$Auto_S$; MS CoT+$Auto_S$;SS CoT+$FIS_S$)\end{tabular}                                                               \\ \hline
6                          & \begin{tabular}[c]{@{}l@{}}zero-shot ( code representation; w/. schema explanation)\\ one-shot ( SS CoT+$Auto_S$; MS CoT+$Auto_S$;SS CoT+$FIS_S$) \\three-shot( SS CoT+$Auto_S$)\end{tabular}                              \\ \hline
7                          & \begin{tabular}[c]{@{}l@{}}zero-shot ( code representation; w/. schema explanation)\\ one-shot ( SS CoT+$Auto_S$; MS CoT+$Auto_S$;SS CoT+$FIS_S$)\\ three-shot( SS CoT+$Auto_S$; MS CoT+$Auto_S$;)\end{tabular}          \\ \hline
8                          & \begin{tabular}[c]{@{}l@{}}zero-shot ( code representation; w/. schema explanation)\\ one-shot ( SS CoT+$Auto_S$; MS CoT+$Auto_S$;SS CoT+$FIS_S$)\\ three-shot( SS CoT+$Auto_S$; MS CoT+$Auto_S$; SS CoT+$FIS_S$)\end{tabular} \\ \hline
\end{tabular}
\label{table: ensemble_settings}
\end{table}
\subsection{Error analysis}\label{appendix-error}

Errors are categorized into five types: schema-linking, join, group-by, nested, and other.
\paragraph{Schema-linking}
This category includes failed queries where the model did not correctly identify column names, table names, or entities mentioned in the questions. It also contains errors where two tables are joined using incorrect foreign keys. Compared to its original definition in Pourreza's study, we have recategorized it to include incorrect join operations using incorrect foreign keys, previously categorized as join errors. This error category is divided into two subcategories: incorrect identification of columns and incorrect identification of tables.

\paragraph{Join}
With the help of the new definition of schema-linking error, our focus about join errors is not on whether the incorrect join is caused by incorrect table or column identification. Instead, we are concerned with whether the generated query includes unnecessary joins that incorrectly narrow down the returned responses or omits essential join operations, leading to incorrect responses.

\paragraph{Group-by}
We find that one scenario defined in the previous definition, where the generated response did not recognize the need for grouping, no longer exists. Therefore, we now only have a subcategory for cases where the response uses the GROUP BY clause on an incorrect column.

\paragraph{Nested}
The original definition of this type of error was that the model did not recognize the nested structure or was unable to detect the correct nesting or set operation. However, in our experiments, we find that the more frequent issue is the GPT engine using nested structures incorrectly — either the nesting is unnecessary and could be replaced by a join operation, or the nested subquery is incorrect. Therefore, we redefine this error type as using nested queries incorrectly, as well as incorrectly using set operations.

\paragraph{Other}

This category includes cases that do not fit under any of the previously mentioned categories. In addition to the misuse of clauses such as ORDER BY, SELECT, and COUNT, and errors in string matching operations, which directly impact the query output, we also consider whether the reasoning engine thoroughly understood the question. In some cases, the engine generates a query that is less related to the question itself, resulting in errors.

\end{document}